%% file: StarVLA.tex
\documentclass{article}

\usepackage{graphicx}
\usepackage{booktabs}
\usepackage{multirow}
\usepackage{algorithm}
\usepackage{algorithmic}
\usepackage{amsmath}
\usepackage{tabularx}
\usepackage{textcomp}

\usepackage[dvipsnames]{xcolor}
\usepackage{colortbl}
\usepackage{array}
\usepackage{rotating}
\usepackage{natbib}

\usepackage{graphicx}
\usepackage{subcaption}
\usepackage{CJKutf8}
\usepackage{enumitem}

\usepackage{pifont}
\usepackage{xcolor}
\usepackage{adjustbox}

\newcommand{\cmark}{\textcolor{green}{\ding{51}}}
\newcommand{\xmark}{\textcolor{red}{\ding{55}}}

\usepackage{booktabs}
\usepackage{array}
\usepackage{adjustbox}
\usepackage{makecell}

\newcommand{\mypara}[1]{\smallskip\noindent\textbf{#1}}

\usepackage[final]{neurips_2024}
\reportdate{April 2026}
\reportprojectpage{\url{https://starvla.github.io}}

\usepackage[utf8]{inputenc}
\usepackage[T1]{fontenc}
\usepackage{hyperref}
\usepackage{url}
\usepackage{booktabs}
\usepackage{amsfonts}
\usepackage{nicefrac}
\usepackage{microtype}
\usepackage{xcolor}

\usepackage{tocloft}

\usepackage{booktabs}
\usepackage{multirow}
\usepackage{graphicx}
\usepackage[table]{xcolor}

\title{StarVLA: A Lego-like Codebase for Vision-Language-Action Model Developing}

\author{
\href{https://starvla.github.io/contributors/}{StarVLA Community} \& \href{https://vni.ust.hk/}{Von Neumann Institute, HKUST}
}

\begin{document}

\maketitle

\input{Sections/A0-Abstract}

\input{Sections/A1-Introduation}

\input{Sections/A2-Various-Frameworks}

\input{Sections/A4-Various-Training}

\input{Sections/A3-Various-Benchmark}

\input{Sections/A5-Experiments}

\input{Sections/A6-All-Bench-in-One}

\input{Sections/A7-Efficiency-Report}

\clearpage

\tableofcontents
\clearpage

{\small

\bibliographystyle{apalike}
\bibliography{egbib}
}

\clearpage
\clearpage

\input{Sections/Tables/author_list}

\end{document}

%% file: Sections/A0-Abstract.tex
\begin{abstract}
Building generalist embodied agents requires integrating perception, language understanding, and action, which are core capabilities addressed by Vision-Language-Action (VLA) approaches based on multimodal foundation models, including recent advances in vision-language models and world models. Despite rapid progress, VLA methods remain fragmented across incompatible architectures, codebases, and evaluation protocols, hindering principled comparison and reproducibility. We present \textbf{StarVLA}, an open-source codebase for VLA research. StarVLA addresses these challenges in three aspects. First, it provides a modular backbone--action-head architecture that supports both VLM backbones (e.g., Qwen-VL) and world-model backbones (e.g., Cosmos) alongside four representative action-decoding paradigms, all under a shared abstraction in which backbone and action head can each be swapped independently. Second, it provides reusable training strategies, including cross-embodiment learning and multimodal co-training, that apply consistently across supported paradigms. Third, it integrates major benchmarks, including LIBERO, SimplerEnv, RoboTwin~2.0, RoboCasa-GR1, and BEHAVIOR-1K, through a unified evaluation interface that supports both simulation and real-robot deployment. StarVLA also ships simple, fully reproducible single-benchmark training recipes that, despite minimal data engineering, already match or surpass prior methods on multiple benchmarks with both VLM and world-model backbones. To our best knowledge, StarVLA is one of the most comprehensive open-source VLA frameworks available, and we expect it to lower the barrier for reproducing existing methods and prototyping new ones. \textbf{\textit{StarVLA is being actively maintained and expanded; we will update this report as the project evolves.}}
The code and documentation are available at \href{https://github.com/starVLA/starVLA}{github.com/starVLA/starVLA}. 
\end{abstract}

%% file: Sections/A1-Introduation.tex
\section{Introduction}

Embodied AI is advancing toward general-purpose agents that integrate perception, language understanding, and action in the physical world, driven in part by recent breakthroughs in large foundation models~\citep{GPT-4, qwen3vl, gao2025seedance}. Vision-Language-Action (VLA) models have emerged as a dominant paradigm for this goal, with a diverse range of design choices. Existing approaches can be broadly grouped into two families: \emph{VLM-based methods}, which repurpose the language model’s representational capacity for action decoding, and \emph{world-model-based methods}, which employ generative architectures to jointly model action distributions and future observations. While both directions have shown strong promise, they are often developed in isolation, with different codebases, interface assumptions, and evaluation protocols, making it challenging to systematically compare them and understand the trade-offs between different design choices.

\paragraph{Fragmentation hinders systematic exploration.} Despite this progress, VLA research remains hindered by fragmentation at multiple levels. At the \textit{architecture} level, existing approaches~\citep{openvla-oft, RT-1, RT-2, bjorck2025gr00t, pi_0, pi_05, wu2026pragmatic, lingBotVA} adopt diverse action-decoding designs, from VLM-native methods (autoregressive tokenization, parallel regression) to generative-model-based methods (diffusion, flow matching), making systematic comparison across paradigm families difficult. At the \textit{system} level, methods are released with tightly coupled assumptions on model architecture, data processing, and training pipelines, limiting component reuse across projects. At the \textit{evaluation} level, results are reported on disjoint subsets of benchmarks with inconsistent protocols, making fair comparison infeasible. Together, these issues create a ``Tower of Babel'' for VLA research, where ideas are difficult to compare, reproduce, or recombine. We attribute this fragmentation to the \textit{lack of a unified abstraction for VLA systems}. Existing codebases~\citep{bjorck2025gr00t, pi_0} are largely method-specific and do not support (i) modular composition across different action-decoding paradigms, (ii) reusable training across heterogeneous data sources, or (iii) standardized evaluation and deployment across benchmarks and embodiments.

\paragraph{StarVLA: a unified platform for exploring embodied intelligence.} We introduce \textbf{StarVLA}, an open-source research platform that brings VLM-based and world-model-based VLA paradigms into a unified modular framework. The core design is a \emph{backbone–action-head decomposition}, where a shared vision-language backbone encodes the scene and instruction, and a pluggable action head maps the resulting representation to motor commands. This formulation is flexible enough to support a wide range of existing approaches, including autoregressive tokenization, parallel regression, flow-matching denoising, and dual-system reasoning, with re-implementations that match or in some cases exceed reported performance. In practice, StarVLA provides three core capabilities:

\begin{itemize}[leftmargin=0.15in]
    \item \textbf{Unified VLA frameworks:} StarVLA implements four representative paradigms under the shared backbone--action-head abstraction (Section~\ref{sec:xFrameworks}): StarVLA-FAST (autoregressive tokenization), StarVLA-OFT (parallel regression), StarVLA-$\pi$ (flow-matching denoising), and StarVLA-GR00T (dual-system reasoning). Crucially, both VLM backbones (e.g., Qwen3-VL) and world-model backbones (e.g., Cosmos-Predict2) are supported as drop-in alternatives, enabling direct comparison between VLM-based and world-model-based research paths under identical training and evaluation conditions. All variants share the same data interface and downstream infrastructure; only the backbone or the action head differs, enabling researchers to isolate the effect of any single design choice while holding all others constant.

    \item \textbf{Flexible training recipes:} StarVLA treats cross-embodiment learning and multimodal co-training as reusable, paradigm-agnostic configurations rather than method-specific add-ons. The same training infrastructure supports supervised action learning, co-training with web-scale vision-language data to preserve multimodal reasoning, and cross-embodiment pretraining across heterogeneous robot datasets. Every recipe applies uniformly to all supported paradigms, making it straightforward to study how training strategies interact with different architectural choices.
    
    \item \textbf{Broad benchmark integration:} StarVLA integrates five mainstream benchmarks (LIBERO, SimplerEnv, RoboTwin~2.0, RoboCasa-GR1, and BEHAVIOR-1K) through a unified server-client testing interface, enabling controlled comparison across environments and embodiments. For each benchmark, we provide simple, fully reproducible training recipes with minimal data engineering that already achieve competitive or state-of-the-art performance under both VLM and world-model backbones, lowering the barrier for the community to build upon. The same interface supports both simulation evaluation and real-robot deployment without code changes, closing the gap between research exploration and practical deployment.
\end{itemize}

To position StarVLA within the existing ecosystem, we compare it with representative open-source VLA systems across key capabilities in Table~\ref{tab:comparison}. To the best of our knowledge, StarVLA is the first platform to bring these capabilities together within a unified interface. Leveraging the controlled comparisons enabled by this framework, StarVLA achieves competitive, and in some cases state-of-the-art, performance across multiple benchmarks with both VLM and world-model backbones, demonstrating that the platform serves not only as a research toolkit but also as a provider of strong, easy-to-reproduce baselines.

\input{Sections/Tables/comparison}

\paragraph{A generalized VLA perspective.}
Beyond its engineering utility, StarVLA also suggests a broader perspective on unifying diverse VLA approaches. Empirically, we find that a single backbone–action-head abstraction can accommodate VLM-based decoding, generative-model-based decoding, and dual-system architectures, all within a shared data pipeline, training loop, and evaluation protocol. This observation indicates that VLM-based and world-model-based methods may be better understood not as fundamentally distinct paradigms, but as variations within a common structural framework, differing primarily in the form of auxiliary learning signals (e.g., language-aligned reasoning or future observation prediction). We refer to this as the \emph{generalized VLA} perspective. Rather than a purely conceptual viewpoint, it arises from the practical unification enabled by StarVLA: when differences in infrastructure are minimized, underlying commonalities become more apparent. We hope this perspective encourages more systematic and cumulative exploration of robotic foundation models.

%% file: Sections/Tables/comparison.tex
\begin{table}[t]
  \centering
  \caption{%
    Comparison of representative open-source VLA systems.
    \textbf{Modular Action Heads}: action heads are plug-and-play on a shared backbone.
    \textbf{Modular VLM}: supports swapping the VLM backbone.
    \textbf{Modular WA}: supports world model as VL backbone.
    \textbf{Mixture DS}: built-in mixture dataloader for heterogeneous data sources.
    \textbf{Open-Source MM Co-train}: open-source multimodal co-training support.
    \textbf{Open-Source X-Emb. Co-train}: open-source cross-embodiment co-training support.
    \textbf{\#Sim Bench}: number of integrated simulation benchmarks with evaluation code.
    \textbf{Multi-Bench Co-train}: joint all benchmarks into one model.
  }
  \label{tab:comparison}
  \begin{adjustbox}{width=\linewidth}
  \begin{tabular}{l c c c c c c c c}
    \toprule
    \textbf{Framework}
      & \makecell[c]{\textbf{Modular Action}\\\textbf{Heads}}
      & \makecell[c]{\textbf{Modular}\\\textbf{VLM}}
      & \makecell[c]{\textbf{Modular}\\\textbf{WA}}
      & \makecell[c]{\textbf{Mixture}\\\textbf{DS}}
      & \makecell[c]{\textbf{Open-Source}\\\textbf{MM Co-train}}
      & \makecell[c]{\textbf{Open-Source}\\\textbf{X-Emb. Co-train}}
      & \makecell[c]{\textbf{\#\textbf{Bench}}}
      & \makecell[c]{\textbf{Multi-Bench}\\\textbf{Co-train}} \\
    \midrule
    OpenPI~\cite{pi_05}
      & \xmark & \xmark & \xmark & \xmark & \xmark & \xmark & 2 & \xmark \\
    Isaac-GR00T~\cite{bjorck2025gr00t}
      & \xmark & \xmark & \xmark & \cmark & \xmark & \cmark & 6 & \xmark \\
    OpenVLA-OFT~\cite{openvla-oft}
      & \xmark & \xmark & \xmark & \xmark & \xmark & \cmark & 1 & \xmark \\
    Dexbotic~\cite{dexbotic}
      & \xmark & \cmark & \xmark & \xmark & \cmark & \xmark & 5 & \xmark \\
    X-VLA~\cite{zheng2025x}
      & \xmark & \xmark & \xmark & \cmark & \xmark & \cmark & 5 & \xmark \\
    \midrule
    \textbf{StarVLA (Ours)}
      & \cmark & \cmark & \cmark & \cmark & \cmark & \cmark & 7 & \cmark \\
    \bottomrule
  \end{tabular}
  \end{adjustbox}
\end{table}

%% file: Sections/A2-Various-Frameworks.tex
\section{Unified Framework for VLA Systems}
\label{sec:xFrameworks}

The rapid evolution of Vision-Language-Action (VLA) models has led to a wide range of heterogeneous designs, with varying preprocessing pipelines, model boundaries, and inference assumptions. While this diversity enables rapid exploration, it often hinders reproducibility and makes fair comparison difficult.  To address this, StarVLA adopts a \emph{unified framework abstraction} at the system level: each method is implemented as a modular component with explicit training and inference interfaces, such that algorithmic differences are isolated to a minimal set of interchangeable modules.

\paragraph{Abstraction of VLA systems.} Beyond the system-level abstraction, we introduce a unified \emph{policy-centric formulation} of VLA models. Prior work often distinguishes between VLM-based policies (VLA) and world-model-based approaches (WAM); here, we place them under a common perspective centered on action generation.

\begin{figure}[h]
    \centering
    \includegraphics[width=1\linewidth]{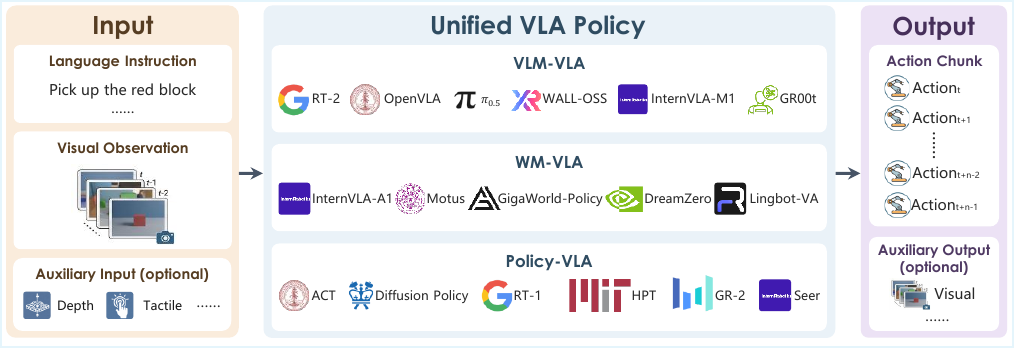}
    \caption{Conceptual view of the unified VLA formulation adopted in StarVLA. A policy $\pi$ maps visual observations and a language instruction to a future action chunk. The training objective decomposes as $\mathcal{L} = \mathcal{L}_{\mathrm{action}} + \mathcal{L}_{\mathrm{aux}}$, where different model families correspond to different forms of $\mathcal{L}_{\mathrm{aux}}$.}
    \label{fig:vla-form}
\end{figure}

As illustrated in Fig.~\ref{fig:vla-form}, we model a VLA system as a policy that maps vision-language (VL) inputs to future action (A) sequences and optional auxiliary outputs:
\begin{equation}
    \pi(\mathbf{a}_{t:t+k}, \mathbf{y}_{\mathrm{aux}} \mid \mathbf{x}_{\leq t}, \ell),
\end{equation}
where:
\begin{itemize}[leftmargin=0.2in]
    \item $\mathbf{x}_{\leq t} = \{o^{\mathrm{vis}}_{\leq t}, o^{\mathrm{depth}}_{\leq t}, o^{\mathrm{tactile}}_{\leq t}, \ldots\}$ denotes the multimodal observation history up to time $t$, which may include visual observations, depth maps, tactile feedback, proprioceptive states, or other sensor modalities;
    \item $\ell$ is the language instruction describing the task;
    \item $\mathbf{a}_{t:t+k}$ represents the predicted $k$-step action chunk from time $t$ to $t+k$;
    \item $\mathbf{y}_{\mathrm{aux}}$ denotes optional auxiliary outputs over the future horizon, such as predicted future visual observations $o^{\mathrm{vis}}_{t+1:t+k}$, intermediate language reasoning or sub-goal descriptions $\ell_{\mathrm{plan}}$, or other modality predictions.
\end{itemize}

This formulation abstracts away intermediate representations and implicitly marginalizes over latent predictions when present, allowing both direct policies and model-based approaches to be expressed within a common interface.

The training objective takes the general form
\begin{equation}
    \mathcal{L} \;=\; \mathcal{L}_{\mathrm{action}} \;+\; \mathcal{L}_{\mathrm{aux}},
\end{equation}
where $\mathcal{L}_{\mathrm{action}}$ supervises the predicted actions, and $\mathcal{L}_{\mathrm{aux}}$ serves as an inductive bias that shapes the learned representation. Different VLA paradigms can then be interpreted as instantiations of this formulation with distinct learning signals:
\begin{itemize}[leftmargin=0.2in]
    \item \textbf{Direct VLA Modeling} sets $\mathcal{L}_{\mathrm{aux}}=0$, optimizing actions alone.
    \item \textbf{VLM-based VLA} introduces language-aligned auxiliary objectives, such as sub-task planning, spatial grounding, or structured reasoning supervision, requiring the model to generate language tokens as auxiliary outputs.
    \item \textbf{WM-based VLA} incorporates future observation prediction (e.g., images or videos), either as an auxiliary objective or as an implicit latent structure that supports action generation, where the model must predict visual states as auxiliary outputs.
\end{itemize}

Under this view, seemingly different paradigms such as VLM-based, world-model-based, and direct policies can be understood as variations of a shared policy formulation with different inductive biases. This perspective simplifies comparison while remaining compatible with both step-wise execution and multi-step open-loop control.

\subsection{Background: VL Foundation Models for Embodied Intelligence}

Embodied agents interact continuously with the physical world, where vision serves as the primary modality for perceiving scene structure, object identity, spatial relations, and interaction affordances.

\noindent\textbf{Vision-language foundation models.} This central role of vision has driven advances in visual representation learning, from supervised models such as ResNet~\citep{he2016resnet} and Vision Transformers~\citep{dosovitskiy2021vit} to scalable self-supervised approaches~\citep{oquab2023dinov2} and video pretraining that captures temporal structure. Building on these backbones, language-aligned pretraining~\citep{radford2021clip, zhai2023siglip} enables shared vision–language representations, while promptable systems such as SAM~\citep{kirillov2023sam, liu2023groundingdino} extend open-world perception. Together with instruction-tuned VLMs~\citep{liu2023llava, chen2023palix, karamcheti2024prismatic, Qwen2.5-VL, hurst2024gpt, georgiev2024gemini15} and generative video models~\citep{gao2025seedance, deepmind2025veo3}, these advances significantly enhance perceptual grounding. However, perception alone is insufficient: embodied agents must also reason over language-conditioned goals and predict environment dynamics under action. Existing models are not inherently designed for action generation or visuomotor control~\citep{act, diffusion, 3ddiffusionpolicy}.

\noindent\textbf{Vision-language modeling for robotic perception and reasoning.} Vision-language pretraining grounds perception in language, providing a scalable interface for task specification and high-level reasoning~\citep{radford2021clip, zhai2023siglip}. Extending this paradigm, Vision-Language-Action (VLA) models incorporate action supervision to unify perception, language, and control~\citep{RT-1, RT-2, openvla, pi_0, intelligence2025pi05, bjorck2025gr00t}. Early works~\citep{nair2022r3m, xiao2022mvp} show that strong visual priors improve control, while VLA models directly map observations and instructions to actions via behavior cloning or policy learning. By transferring large-scale semantic knowledge into control, they improve instruction following and cross-task generalization, often outperforming prior robotic policies~\citep{act, chi2024diffusionpolicyvisuomotorpolicy, 3ddiffusionpolicy}. Recent work extends these models to humanoid loco-manipulation~\citep{psi0}.
To preserve reasoning capabilities, subsequent approaches explore multimodal co-training~\citep{pi_05KI, ye2026st4vla, chen2025internvla, zeng2024molmoact, instructvla, zhou2025chatvla}, while others scale teleoperated datasets~\citep{open_x_embodiment, khazatsky2024droid, robomind, agibot, Bridge_data, Manipulate-anything}. However, these datasets remain limited in task, language, and scene diversity~\citep{hirobot, umi}, motivating portable data collection~\citep{generalistai, liu2024rdt, umi}. Meanwhile, tightly coupled pipelines hinder reproducibility, modularity, and scalability, highlighting the need for unified frameworks. 

\noindent\textbf{Video-based world model for robotic dynamics and interaction.} 
Orthogonal to language-based scaling, video-based world models learn physical dynamics via visual prediction. Video captures motion, contact, and causality more effectively than static data. Early methods augment VLA policies with predictive latent modeling~\citep{flare, bjorck2025gr00t, ye2025lapa}, while large-scale video pretraining enables planning with minimal robot data~\citep{vjepa2, dreamGen}. Later work treats video as a primary policy substrate, either unifying policy, simulation, and evaluation or decoupling planning from control~\citep{du2023unipi, ko2024avdc, mimicVideo, lvp}.
Action-conditioned world models further support policy evaluation and improvement: imagined rollouts achieve strong performance~\citep{dayDreamer}, while recent systems enable counterfactual replay and safety evaluation~\citep{1xWM, veorobotics2025}. Other approaches use controllable world models for trajectory generation, reinforcement learning, or scalable data synthesis~\citep{ctrlWorld, woVR, gigaWorld0, gigaBrain05M, swirl}.
Recent work emphasizes causal consistency, controllability, and closed-loop efficiency by integrating action and value prediction into pretrained video models or jointly learning dynamics and control~\citep{cosmosPolicy, lingBotVA, cai2026internvla, dreamDojo, emma, motus}. Some approaches formulate joint video–action prediction as policy learning or analyze gains from test-time imagination versus co-training~\citep{dreamZero, fastWAM}. Additionally, human video provides scalable motion priors, with egocentric pipelines enabling transferable behaviors across tasks and embodiments~\citep{egoDex, egoVLA, egoScale}.

Above vision-language pretraining and video-based world modeling scale embodied intelligence along complementary but largely fragmented axes, motivating StarVLA's design choice to separate what should vary across methods from what should remain stable across training, evaluation, and deployment.

\subsection{Building VLA Frameworks on VL Foundation Models}

While the foundation models surveyed above provide powerful visual-linguistic representations, they are not natively designed for action generation. A key design goal of StarVLA is to make these VL foundation models \emph{VLA-ready}: we provide a unified I/O interface contract and a compositional architecture that allow diverse action decoding strategies to be flexibly composed on top of the same VL backbone.

\paragraph{Unified I/O Interface.}
All framework modules in StarVLA inherit from a common base class and expose two methods that share a unified \emph{input/output (I/O) interface}: both training and inference consume raw, environment-level observations identical to what the robot receives at deployment time.
\begin{itemize}[leftmargin=0.2in]
    \item \texttt{forward(\{raw images, str, ...\})} $\rightarrow$ \texttt{\{raw images, str, ...\}}: the training entry point. It receives a batch of raw samples, each containing multi-view RGB images, a natural-language instruction, and an action chunk, and returns a loss dictionary.
    \item \texttt{predict\_action(\{raw images, str, ...\})} $\rightarrow$ \texttt{\{normalized\_actions, ...\}}: the inference entry point. It accepts the same observation format (minus ground-truth actions) and returns predicted action chunks.
\end{itemize}
By deliberately adopting this unified I/O interface, where training inputs mirror real deployment observations rather than relying on heavily preprocessed dataloader tensors, we minimize train/test distribution mismatch, a common source of silent performance degradation in VLA systems.

This design choice reflects a deeper invariant of robotic deployment: regardless of how different VL foundation models are pretrained---what tokenization scheme they adopt, how they resize or partition images, or what auxiliary objectives they optimize during pretraining---at inference time every model must ultimately accept the same raw sensor streams that the physical robot provides and produce executable motor commands. The unified I/O interface codifies this \emph{deployment-time invariant} as the system's first-class contract, ensuring that any VL model whose inference path can consume raw observations is immediately compatible with StarVLA, without requiring users to reverse-engineer or replicate model-specific preprocessing pipelines. Crucially, this same invariant-driven principle extends naturally to the internal architecture, as we describe next.

\paragraph{Compositional framework.}
Applying the same principle internally, we decompose every VLA method into two explicitly separated components connected by a standardized representation contract: a \textbf{VL backbone} (e.g., Qwen2.5-VL, z) that consumes raw multimodal observations and exposes hidden-state representations through a common output specification, and a \textbf{pluggable action head} that reads those representations through a corresponding input specification and converts them into motor commands. Each framework assembles itself through the same two-step composition (first loading the backbone, then attaching an action head), with both components configured declaratively via YAML.
Because the outer system boundary (raw observations $\to$ actions) and the inner backbone--head boundary (multimodal inputs $\to$ hidden states $\to$ actions) are both governed by standardized contracts, StarVLA achieves \emph{bidirectional modularity}: backbone and action head can each be replaced independently without affecting the other or any surrounding infrastructure.

This modularity provides flexibility across different stages of VLA development. \textbf{For researchers}, it supports rapid experimentation in multiple directions. New action decoding paradigms can be prototyped by implementing and registering an action-head module, while new vision-language backbones—such as instruction-tuned VLMs (e.g., Qwen2.5-VL~\citep{Qwen2.5-VL}, InternVL~\citep{cai2026internvla}) or video-native models (e.g., Cosmos~\citep{cosmosPolicy})—can be integrated through a lightweight adapter that conforms to a shared representation interface. Once integrated, these backbones can be evaluated across different action heads without requiring per-method modification. \textbf{For training infrastructure}, the standardized interfaces allow much of the upstream and downstream stack (e.g., training pipelines, benchmark harnesses, and deployment services) to remain largely \emph{backbone- and action-head-agnostic}, reducing the need for method-specific code paths as new paradigms or models are introduced. \textbf{For deployment}, switching between different backbones or action paradigms can be handled through configuration changes, without requiring code-level modifications.

\subsection{Representative VLA Instantiations}
Under this unified abstraction, we implement four paradigms spanning the major action decoding families in the current VLA literature, as illustrated in Fig.~\ref{fig:a2-frameworks}. All variants share the same VL backbone, the same base class, and the same \texttt{forward}/\texttt{predict\_action} contract, differing only in \emph{how they extract actions from the backbone's representations}:

\begin{figure}[h]
	\centering
	\includegraphics[width=\linewidth]{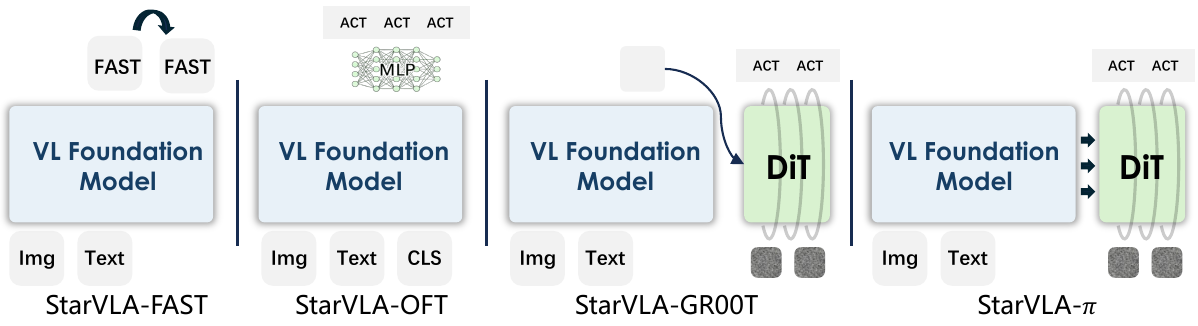}
	\caption{Overview of four representative approaches for adapting Vision-Language Models into Vision-Language-Action frameworks in StarVLA (FAST, OFT, $\pi$, and GR00T) under a unified interface. 
    }
	\label{fig:a2-frameworks}
\end{figure}

\begin{itemize}[leftmargin=0.2in]
	\item \textbf{StarVLA-FAST} ($\pi_{\text{fast}}$): Appends a FAST tokenizer~\citep{pertsch2025fast} to the VL backbone and \emph{autoregressively} generates discrete action tokens via next-token prediction, using the LLM's own vocabulary space.
	
	\item \textbf{StarVLA-OFT:} Attaches a lightweight MLP that reads the hidden states of predefined action tokens and \emph{regresses} continuous actions in parallel (L1 loss), following OpenVLA-OFT~\citep{openvla-oft}---the simplest form of pluggable head.i
	
	\item \textbf{StarVLA-$\pi$} ($\pi_0$): Integrates a layer-wise cross-DiT flow-matching action expert, conditioned on multi-layer VL hidden states via cross-attention, and predicts continuous actions through iterative \emph{denoising}, following $\pi_0$~\citep{pi_0}.
	
	\item \textbf{StarVLA-GR00T:} Adopts a dual-system design where the VL backbone serves as \emph{System~2} (slow reasoning) and a DiT-based flow-matching module serves as \emph{System~1} (fast action generation), consistent with GR00T N1.5~\citep{bjorck2025gr00t}. This variant demonstrates that even fundamentally different inference-time compute patterns can coexist under the same interface.
\end{itemize}

This spectrum, from VLM-native decoding (autoregressive tokenization, parallel regression) to generative-model-based decoding shared with world-model architectures (iterative flow-matching denoising, dual-system reasoning), shows that the proposed compositional architecture and unified interface are broadly applicable. Adding further paradigms requires only implementing and registering a new action head; the backbone, training loop, and evaluation pipeline remain unchanged.

%% file: Sections/A4-Various-Training.tex
\section{Unified System Pipeline for Model Training and Testing}
\begin{figure}[h]
	\centering
    \includegraphics[width=\linewidth]{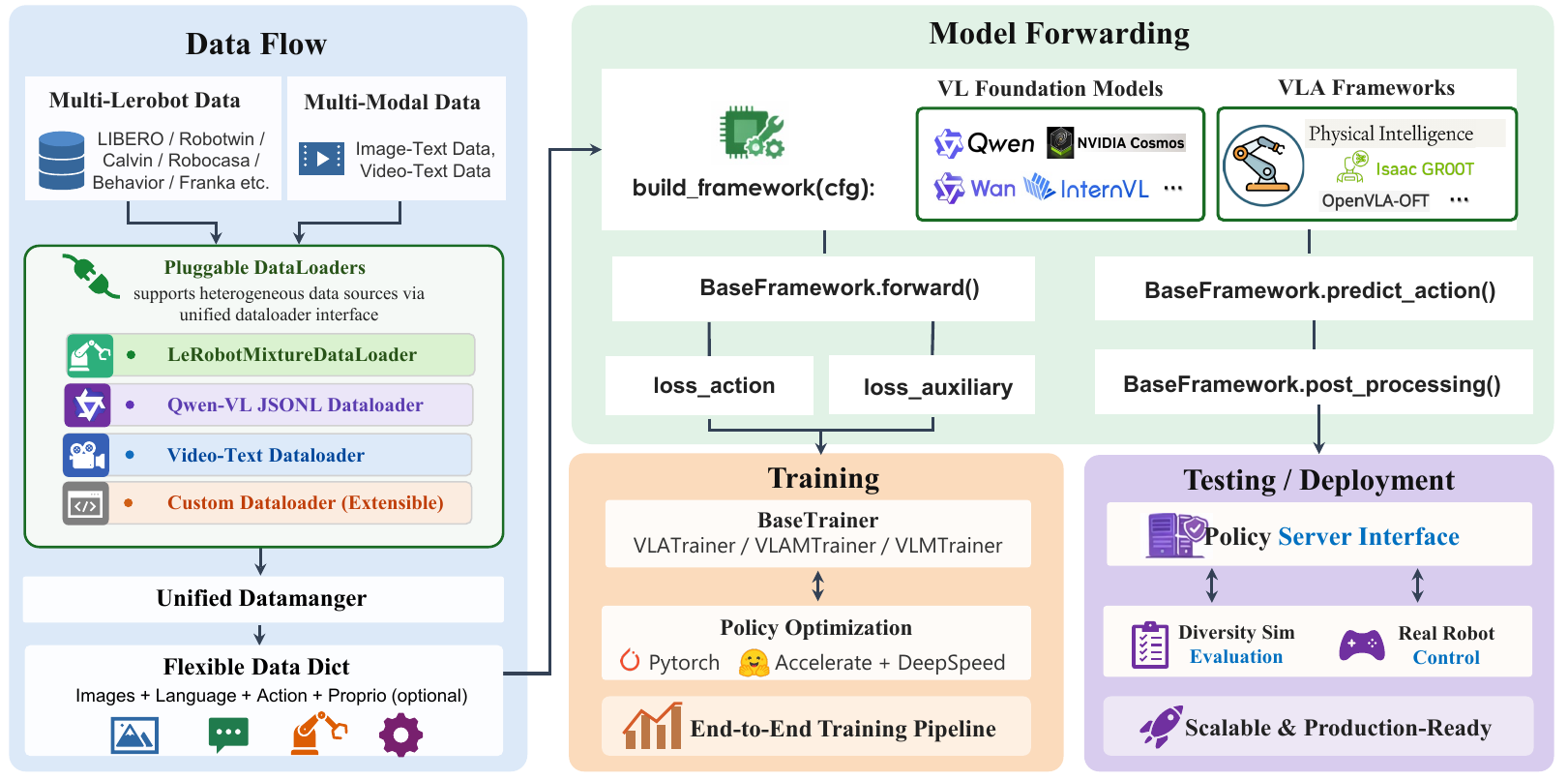}
	\caption{Overview of the StarVLA framework. We present a unified and modular pipeline that connects heterogeneous data sources, pluggable dataloaders, and flexible data representations with a standardized model forwarding interface. The framework supports diverse vision-language foundation models and VLA architectures, enabling end-to-end training and deployment.}
	\label{fig:training-overview}
\end{figure}

The StarVLA codebase supports several practical training regimes for VLA policies, ranging from standard supervised fine-tuning (SFT) on downstream robot datasets to multi-objective co-training with vision--language (VLM) web data and cross-embodiment co-training on mixed robot embodiments. All training pipelines are implemented in explicit PyTorch loops built on Accelerate + DeepSpeed for distributed execution, while preserving a unified YAML configuration interface across methods.
Figure~\ref{fig:training-overview} summarizes the supported training modes and how data streams connect to the unified model framework.

\subsection{Training Paradigms}

\subsubsection{Supervised Learning    for Behavior Cloning}
The most direct training mode is robot-only supervised learning, where the policy is trained to predict continuous actions from observations and language instructions.
In our codebase, this training path is implemented in \texttt{starVLA/training/train\_starvla.py}.
The objective is the action modeling loss returned by the framework \texttt{forward()} method (e.g., \texttt{action\_loss} in the output dict).

\paragraph{Optimization setup.}
We support (i) full-parameter fine-tuning and (ii) selective freezing of submodules via \texttt{trainer.freeze\_modules} (comma-separated module paths).
To stabilize training across heterogeneous components, the optimizer can use multiple parameter groups with different learning rates (e.g., separate LR for \texttt{qwen\_vl\_interface} and the action model) configured by \texttt{trainer.learning\_rate}.
Training uses bfloat16 autocast, gradient accumulation, gradient clipping, and a cosine schedule with a minimum learning rate.

\subsubsection{Multi-Objective Co-Training for Embodied Reasoning}
Robot-only SFT can over-specialize the VLM backbone to a narrow instruction distribution.
To preserve general-purpose visual reasoning and language grounding while learning action prediction, StarVLA supports a co-training regime that interleaves robot action learning with a VLM loss on multimodal web data.
This mode is implemented in \texttt{starVLA/training/train\_starvla\_cotrain.py}.

\paragraph{Dual-loader multi-objective training scheme.}
Co-training uses two dataloaders (VLA and VLM) and performs two forward/backward passes per optimization step:
(i) a VLA forward pass through the framework \texttt{forward()} to obtain \texttt{action\_loss}, and
(ii) a VLM forward pass through \texttt{qwen\_vl\_interface} to obtain the language modeling loss.
The VLM loss is scaled by \texttt{trainer.loss\_scale.vlm} in config, enabling a controlled trade-off between action learning and VLM capability retention.

\subsubsection{Cross-Embodiment Co-Training with Robot Data Mixtures}
To support cross-embodiment generalization, the codebase provides a unified LeRobot mixture dataset interface that allows training on heterogeneous robot datasets with different embodiments, action conventions, and camera setups.
In config, users select a named mixture through \texttt{datasets.vla\_data.data\_mix}, which maps to a list of (dataset name, sampling weight, robot type) tuples.
At runtime, the mixture is materialized as a \texttt{LeRobotMixtureDataset}, which samples trajectories across datasets according to the specified weights and tracks embodiment tags based on robot type.
This design makes ``cross-embodiment pretraining'' an operational configuration choice, rather than a bespoke training script.

\subsubsection{Reinforcement Learning Fine-Tuning}
Beyond supervised and co-training regimes, we plan to support reinforcement learning (RL) fine-tuning as an extension of the same framework abstraction, collaborating with the RLinf project (\url{https://github.com/RLinf/RLinf}).
At the time of writing, RL fine-tuning is an ongoing integration effort; the current public codebase focuses on supervised and co-training pipelines to build up a strong robotic foundation model.

\subsection{Evaluation and Deployment}

\subsubsection{Unified Server-Client Evaluation Across Benchmarks}

StarVLA adopts a thin server--client testing abstraction so that benchmark-side evaluation code remains close to the official implementations, while model-side inference is standardized. In practice, a checkpoint is loaded by \texttt{baseframework.from\_pretrained()} and hosted as a lightweight WebSocket policy server in the StarVLA runtime environment. The benchmark evaluator, which may live in a different conda environment with its own simulator dependencies, interacts with the model through a small client wrapper rather than importing framework code directly. This decoupling is particularly useful for benchmarks such as LIBERO, SimplerEnv, and RoboTwin, whose official evaluators each carry different dependency stacks and control loops.

\paragraph{Inference interface.}
All framework variants expose the same inference entry point, \texttt{Framework.predict\_action()}, and the server forwards incoming payload dictionaries to this method with minimal routing logic. The benchmark-side client packages observations into a single dictionary, typically containing \texttt{image} (single- or multi-view RGB observations), \texttt{lang} (task instruction), and optional fields such as \texttt{state}, timestamps, or episode metadata. The payload is serialized with \texttt{msgpack} and sent to the policy server, which returns a dictionary containing model outputs such as \texttt{normalized\_actions}. Because the communication contract is action-head-agnostic, switching from OFT to FAST, $\pi$, or GR00T does not require modifying benchmark code.

\paragraph{Benchmark-specific adapters.}
In StarVLA, benchmark differences are isolated in lightweight interface files such as \texttt{model2libero\_interface.py}, \texttt{model2simpler\_interface.py}, and \texttt{model2robotwin\_interface.py}. These adapters translate raw environment observations into the common StarVLA example format and post-process returned actions into the benchmark's native control API. Typical responsibilities include resizing images to the training resolution, reading \texttt{dataset\_statistics.json} from the checkpoint directory for action unnormalization, converting chunked normalized predictions into executable actions, applying action ensembling, and handling benchmark-specific conventions such as sticky grippers or delta/relative-to-absolute action conversion. This design keeps the core policy server benchmark-agnostic while preserving faithful evaluation under each official protocol.

\subsubsection{Deployment on Real Robots}

The same client-server contract also supports real-robot or hosted-benchmark deployment. In this setting, the robot controller plays the role of the benchmark client: it captures camera observations, assembles the same example dictionary used in simulation, queries the remote policy server, and executes the returned action on hardware. As a result, the control loop, safety logic, and device-specific middleware remain outside the StarVLA model runtime, while the model service remains unchanged.

\paragraph{Deployment interface.}
This separation makes deployment much less intrusive. The model stack can stay in a GPU-oriented inference environment, whereas the robot-side process can remain integrated with vendor SDKs, ROS nodes, or hosted evaluation platforms such as RoboChallenge. More importantly, the exact same checkpoint can be reused across simulation and real-robot settings as long as the client provides observations in the agreed dictionary format and applies the appropriate benchmark- or robot-specific post-processing. In this sense, StarVLA treats deployment as a continuation of the same testing paradigm rather than as a separate engineering path.

%% file: Sections/A3-Various-Benchmark.tex
\section{Multiple Benchmark Integration}
\label{sec:a3_benchmarks}

Recent vision-language-action (VLA) research has made rapid progress across a wide range of benchmarks. However, most existing methods are evaluated on only a limited number of environments, and their implementations often differ substantially in preprocessing pipelines, policy interfaces, and evaluation protocols. These inconsistencies hinder fair cross-paper comparison and weaken reproducibility. 

\subsection{Unified Benchmark Integration Interface}

StarVLA aims to provide \emph{simple and reproducible baselines} across a diverse benchmark suite by: 
(i) adhering as closely as possible to the official training and evaluation workflows of each benchmark, with minimal data engineering and environment-specific modifications, and 
(ii) standardizing the policy-side interface. 
Concretely, all StarVLA variants expose a unified lightweight WebSocket service, enabling different benchmark runners to interact with a shared inference endpoint. This design facilitates seamless integration and simplifies scaling to additional benchmarks.

To facilitate reproducibility, StarVLA defines a unified integration interface for benchmark onboarding. Specifically, each benchmark integration is structured around three aligned components:
\textbf{(i)} a checkpoint package containing the saved \texttt{config.yaml} and \texttt{dataset\_statistics.json}, 
\textbf{(ii)} a runnable training entry (YAML configuration and launch script under \texttt{examples/<BENCH>/train\_files/}), and 
\textbf{(iii)} a runnable evaluation workflow that launches a policy server and invokes the official benchmark evaluator (typically under \texttt{examples/<BENCH>/eval\_files/}). 
This design ensures that benchmark-specific workflows remain reproducible while maintaining a consistent policy interface across environments.

\subsection{Supported Benchmark Suite}
StarVLA integrates a diverse set of manipulation benchmarks spanning different simulators, embodiments, and protocols, including \textit{LIBERO}, \textit{SimplerEnv}, \textit{RoboTwin 2.0}, \textit{RoboCasa GR1 Tabletop Tasks}, and \textit{BEHAVIOR-1K}. The experiments section reports detailed results and comparisons under each benchmark's official evaluation protocols.

\noindent\textbf{LIBERO.} LIBERO~\citep{Libero} is a widely used benchmark for language-conditioned robot manipulation and lifelong robot learning. It contains 130 manipulation tasks organized into four suites: Spatial, Object, Goal, and Long, each targeting a different form of generalization, including spatial variation, object-centric manipulation, goal-conditioned execution, and long-horizon dependencies. A standard training protocol uses 50 demonstrations per task, resulting in approximately 6.5K trajectories in total. LIBERO provides a standardized evaluation protocol and serves as a comprehensive testbed for instruction following, compositional generalization, and multi-task policy learning.

\noindent\textbf{LIBERO-Plus.} LIBERO-Plus~\citep{libero_plus} is a robustness-oriented benchmark built on top of LIBERO for systematically evaluating the generalization ability of vision-language-action models under distribution shifts. It expands the original benchmark by introducing perturbations over seven factors, including object layout, camera viewpoints, robot initial states, language instructions, lighting, background textures, and sensor noise. The final benchmark is a test-only evaluation set with 10,030 tasks spanning 7 perturbation factors and 21 low-level components.

\noindent\textbf{SimplerEnv.} SimplerEnv~\citep{simpleenv} is a simulation-based evaluation benchmark designed as a scalable proxy for real-world robot evaluation. It provides standardized simulated environments corresponding to common real-robot platforms, including the WidowX (BridgeData V2) and the Google Robot (RT-series) setups. The benchmark defines fixed evaluation protocols such as Visual Matching and Variant Aggregation, as well as standardized success-rate aggregation rules. Although it does not specify a fixed number of tasks or dataset size, it is widely used to evaluate policies trained on real-world data under reproducible simulated conditions, with prior work showing strong correlation between simulated and real-world performance.

\noindent\textbf{RoboCasa-GR1.} RoboCasa-GR1~\citep{robocasa, bjorck2025gr00t} is a tabletop manipulation benchmark built on the RoboCasa simulation framework, commonly used to evaluate humanoid-style manipulation policies. Compared with standard single-arm setups, it introduces more complex embodiments and household interaction scenarios involving articulated objects and multi-stage tasks. The benchmark contains 24 tasks, with approximately 1,000 demonstrations per task, resulting in around 24K trajectories in total.

\noindent\textbf{RoboTwin 2.0.} RoboTwin 2.0~\citep{chen2025robotwin2} is a large-scale benchmark for bimanual robotic manipulation, focusing on dual-arm coordination across diverse scenarios. It contains 50 tasks with two evaluation setups: clean and randomized. Each task includes 50 clean demonstrations together with 500 randomized demonstrations, resulting in approximately 550 trajectories per task and 27.5K trajectories in total. The randomized data is generated via structured domain randomization, including variations in scene clutter, backgrounds, table height, and lighting, providing a challenging testbed for both coordination and robustness. For evaluation, each task is tested for 100 episodes under each setup. In total, this results in 50 tasks $\times$ 2 setups $\times$ 100 episodes, equals to 10,000 evaluation trials.

\noindent\textbf{BEHAVIOR-1K.} BEHAVIOR-1K~\citep{behavior1k} is a large-scale benchmark for human-centered embodied AI, built around everyday activities. It defines 1,000 activities across 50 interactive scenes with more than 9,000 objects, covering environments such as homes, offices, and restaurants. Built on OmniGibson, it supports realistic physics for rigid, deformable, and liquid objects, and emphasizes long-horizon interaction requiring perception, navigation, and manipulation. An active evaluation setting is the BEHAVIOR Challenge, which selects 50 household tasks from the activity set and provides 10,000 teleoperated demonstrations (over 1,200 hours), with 200 demonstrations per task released for training. For evaluation, each task includes 20 additional instances with varying initial conditions, of which 10 are used for reporting, and each instance is evaluated once with a fixed timeout. Performance is measured by the average task success rate across all tasks, with partial credit based on goal completion.

\noindent\textbf{CALVIN.} CALVIN~\citep{mees2022calvin} is a benchmark for long-horizon language-conditioned manipulation, designed to evaluate whether a single policy can execute sequences of natural-language instructions from visual observations. It comprises four environments (A, B, C, and D) and 34 manipulation tasks involving articulated objects and stateful scene elements such as drawers, sliding doors, lights, and switches. The standard evaluation follows the ABC$\rightarrow$D setting, where policies are trained on A--C and tested on D using 1,000 task sequences of length 5. Performance is reported by the average length of successfully completed subtask sequences.

%% file: Sections/A5-Experiments.tex
\section{Single-Benchmark Training Examples}
\label{sec:Basic_experiments}
In this section, we report \emph{single-benchmark SFT results} to establish transparent, reproducible reference points under official evaluation protocols.
To provide the community with the cleanest possible baselines, we deliberately avoid any VLA-specific pretraining (e.g., large-scale robot pretraining mixtures), data augmentation, or online refinement techniques such as DAgger.
Every model is initialized from publicly released VL pretrained weights and fine-tuned exclusively on the benchmark's standard demonstration dataset.
These minimal-assumption results serve as reliable \emph{anchor points} for future research: they make it straightforward to measure the marginal value of additional pretraining data, augmentation strategies, or co-training recipes.

\input{Sections/Benchmarks/A3-LIBERO}

\input{Sections/Benchmarks/A3-SIMPLER}
\input{Sections/Benchmarks/A3-RoboCasa}

\input{Sections/Benchmarks/A3-RoboTwin}

\input{Sections/A5-CoTraining}

%% file: Sections/Benchmarks/A3-LIBERO.tex
\subsection{Results on LIBERO}

LIBERO~\citep{Libero} is a widely used tabletop manipulation benchmark comprising four task suites of increasing difficulty: \textbf{Spatial}, \textbf{Object}, \textbf{Goal}, and \textbf{Long}.
We treat it as the first worked example of our single-benchmark pipeline and walk through every step—data loading, training, and evaluation—so that readers can fully reproduce our numbers.

\paragraph{Training data format.}
To maintain a simple and reproducible baseline, we adopt minimal data engineering and follow the benchmark's native schema.
\begin{itemize}[leftmargin=0.2in]
  \item \texttt{\textbf{Input}: a raw sample dict loaded directly from the LeRobot-format dataset, containing the primary (third-person) RGB view and the wrist-camera RGB view. We do not use proprioceptive state, history stacking, or image augmentation for this baseline.}
  \item \texttt{\textbf{Output}: a continuous end-effector (EEF) control action vector following the LIBERO action definition with action chunking$=8$.}
\end{itemize}

\paragraph{Training setup.}
We train the LIBERO baseline using distributed training with 8 A100 GPUs (via \texttt{accelerate} + DeepSpeed ZeRO-2). Unless otherwise specified, the per-device batch size is 16 and training runs for 100K optimization steps. Checkpoints are saved every 10K steps, with periodic logging and evaluation during training. For transparency and exact reproducibility (full command line, YAML configuration, and environment variables),we provide the complete training scripts under \texttt{examples/LIBERO/train\_files/}. We train a single policy jointly on four LIBERO suites (Spatial, Object, Goal, and {LIBERO-10}) using the corresponding LeRobot-format datasets:
They are available as a public collection at \url{https://huggingface.co/collections/IPEC-COMMUNITY/libero-benchmark-dataset}. 

\paragraph{Evaluation protocol.}
We evaluate on the four suites (Spatial, Object, Goal, and \texttt{LIBERO-Long}) using the official LIBERO evaluation scripts and report success rate. 
We periodically evaluate checkpoints (every 10K steps by default) and report the earliest checkpoint that achieves the best average success rate.
For each suite, we run 10 tasks with 50 episodes per task (500 trials total) and report the mean success rate over all trials. To ensure reproducibility without modifying benchmark logic, we provide the complete evaluation scripts and launch instructions under \texttt{examples/LIBERO/eval\_files/}.

\paragraph{Results and analysis.}
Table~\ref{tab:libero_results} summarizes the LIBERO baseline performance. 
Using only 30K steps ($\sim$10 epochs), StarVLA already matches or surpasses several strong published baselines.
For instance, OpenVLA-OFT trains for 175K steps (223 epochs) to reach 97.1\% average, whereas StarVLA-OFT achieves 96.6\% (Qwen3-VL) and 95.8\% (Cosmos-Predict2-2B) with $6\times$ fewer steps and $23\times$ fewer epochs.
$\pi_0$+FAST and GR00T-N1.5 score 85.5\% and 86.5\% respectively, both considerably below our variants.
Notably, replacing the VL backbone from Qwen3-VL-4B to Cosmos-Predict2-2B yields comparable performance (average $\geq$95.2\% across all action heads), demonstrating that StarVLA generalizes well across different VL backbones.
These comparisons suggest that the StarVLA pipeline is highly data-efficient on LIBERO.


\begin{table}[h!]
  \centering
  \caption{Comparison of different VLA models on LIBERO. 
  We train one policy for all 4 suites. All scores are averaged over 500 trials for each task suite (10 tasks × 50 episodes). }
  \label{tab:libero_results}
  \begin{adjustbox}{width=0.8\linewidth}
  \begin{tabular}{l cc c c c c c}
    \toprule
    \textbf{Model} & \textbf{Steps} & \textbf{Epochs} & \textbf{Spatial} & \textbf{Object} & \textbf{Goal} & \textbf{Long} & \textbf{Avg} \\
    \midrule
    $\pi_0$+FAST~\cite{pertsch2025fast} & - & - & 96.4 & 96.8 & 88.6 & 60.2 & 85.5 \\
    OpenVLA-OFT~\cite{openvla-oft} & 175K & 223 & 97.6 & 98.4 & 97.9 & 94.5 & 97.1 \\
    $\pi_0$ & - & - & 96.8 & 98.8 & 95.8 & 85.2 & 94.1 \\
    GR00T-N1.5~\cite{bjorck2025gr00t} & 20K & 203  & 92.0 & 92.0 & 86.0 & 76.0 & 86.5 \\
    \midrule

     \multicolumn{8}{c}{\textbf{VL = Qwen3-VL-4B}} \\
    StarVLA-FAST & 30K & 9.54 & 97.3 & 97.4 & 96.3 & 90.6 & 95.4 \\
    StarVLA-OFT & 30K & 9.54 & 97.8 & 98.6 & 96.2 & 93.8 & 96.6 \\
    StarVLA-$\pi$ & 30K & 9.54 & 98.8 & 99.6 & 95.8 & 88.4 & 95.7  \\ 
    StarVLA-GR00T & 30K & 9.54 & 97.8 & 98.8 & 97.4 & 92.0 & 96.5 \\

    \midrule
    \multicolumn{8}{c}{\textbf{VL = Cosmos-Predict2-2B}} \\

    StarVLA-OFT & 30K & 9.54  & 98.6 & 97.6 & 95.0 & 91.8 & 95.8 \\
    StarVLA-$\pi$ & 30K & 9.54  & 98.9 & 98.3 & 94.4 & 90.4 & 95.5 \\
    StarVLA-GR00T & 30K & 9.54  & 97.4 & 98.0 & 95.1 & 90.4 & 95.2 \\

    \bottomrule
  \end{tabular}
  \end{adjustbox}
\end{table}

%% file: Sections/Benchmarks/A3-SIMPLER.tex
\subsection{Results on SimplerEnv}


\paragraph{Training setup.}
All models are trained with full-parameter fine-tuning using distributed training on 16 A100 GPUs.
Unless otherwise specified, the per-device batch size is 16 and training runs for 100K optimization steps. Checkpoints are saved every 10K steps, with periodic logging and evaluation during training. For transparency and exact reproducibility (full command line, YAML configuration, and environment variables), we provide the complete training scripts under \texttt{examples/SimplerEnv/train\_files/}. We train SimplerEnv baselines on a merged mixture of Bridge and Fractal datasets in LeRobot format: \url{https://huggingface.co/datasets/IPEC-COMMUNITY/bridge_orig_lerobot} and \url{https://huggingface.co/datasets/IPEC-COMMUNITY/fractal20220817_data_lerobot}. 

\paragraph{Evaluation protocol.}
We evaluate using the official SimplerEnv evaluation workflow and report the task success rate. We present detailed per-task results under two standard SimplerEnv settings: (i) WidowX robot with Visual Matching (VM) in Table~\ref{tab:simplerenv_widowx_full}, and (ii) Google Robot in Table~\ref{tab:simpler-google}. We strictly follow the official protocol for per-task repeats/episodes and success-rate aggregation without modifying benchmark logic.
Since SimplerEnv evaluation can exhibit non-trivial variance, we run each reported setting five times (each time rerunning the full official evaluation) and report the mean success rate. 
To ensure reproducibility without modifying benchmark logic, we provide the complete evaluation scripts and launch instructions under \texttt{examples/SimplerEnv/eval\_files/}.

\paragraph{Results.}
Tables~\ref{tab:simplerenv_widowx_full} and~\ref{tab:simpler-google} summarize the SimplerEnv performance. On WidowX (VM), StarVLA with Qwen3-VL-4B achieves a strong average success rate (up to 65.3\%), while the Cosmos-Predict2-2B backbone also delivers competitive results (up to 61.6\%), confirming that StarVLA generalizes across different VL backbones. Both configurations show consistently high performance on the most structured task and a remaining gap on object placement tasks. On Google Robot, StarVLA is competitive with or better than strong recent baselines under the Visual Matching setting, and remains comparable under Variant Aggregation, suggesting that the policy transfers robustly across standardized simulation evaluation settings.

\begin{table}[ht]
    \centering
    \caption{Detailed results on the SimplerEnv WidowX benchmark (Visual Matching). Steps denote optimization steps; all numbers are success rates (\%).}
    \label{tab:simplerenv_widowx_full}
    \begin{adjustbox}{width=\linewidth}
    \begin{tabular}{l l c c c c c c}
    \toprule
    \multicolumn{1}{c}{\textbf{WidowX Robot}} & \textbf{Method}
      & \textbf{Steps}
      & \makecell[c]{\textbf{Put Spoon} \\ \textbf{on Towel}}
      & \makecell[c]{\textbf{Put Carrot} \\ \textbf{on Plate}}
      & \makecell[c]{\textbf{Stack Green Block} \\ \textbf{on Yellow Block}}
      & \makecell[c]{\textbf{Put Eggplant} \\ \textbf{in Yellow Basket}}
      & \textbf{Average} \\
    \midrule
    \multirow{20}{*}{\makecell[l]{SIMPLERENV \\ Visual Matching}}
      & RT-1-X~\cite{RT-1}  & - & 0.0  & 4.2  & 0.0  & 0.0  & 1.1 \\
      & Octo-Base~\cite{octo}    & - & 15.8 & 12.5 & 0.0  & 41.7 & 17.5 \\
      & Octo-Small~\cite{octo}  & - & 41.7 & 8.2  & 0.0  & 56.7 & 26.7 \\
      & OpenVLA~\cite{openvla}     & - & 4.2  & 0.0  & 0.0  & 12.5 & 4.2 \\
      & {CogACT}~\cite{cogact}              & - & {71.7}   & {50.8} & {15.0} & 67.5 & {51.3} \\
      & {SpatialVLA}~\cite{spatialvla}              & - & 16.7   & 25.0 & 29.2 & \textbf{100.0} & 42.7 \\

    & {$\pi_0$}~\cite{pi_0}              & - & {29.1} & {0.0}  & {16.6} & {62.5} & 27.1 \\
    & $\pi_0$-FAST~\cite{pertsch2025fast} & - & 29.1 & 21.9 & 10.8 & 66.6 & 48.3 \\
    & GR00T N1.5~\cite{bjorck2025gr00t}   & - & {75.3} & {54.3} & \textbf{{57.0}} & 61.3 & {61.9} \\
    & Magma~\cite{magma}  & - & 37.5 & 31.0 & 12.7  & 60.5 & 35.8 \\
    \cmidrule(lr){2-8}


      & \multicolumn{7}{c}{\textbf{VL = Qwen3-VL-4B}} \\
    & \textbf{StarVLA-FAST} & 15K & 18.8 & 31.3 & 4.2 & 71.9 & 31.6 \\
    & \textbf{StarVLA-OFT}  & 65K & \textbf{90.3} & 38.5 & 29.7 & \textbf{100.0} & 64.6 \\
    & \textbf{StarVLA-$\pi$}  & 40K & 78.1 & 46.9 & 30.2  & 88.5 & 60.9  \\
    & \textbf{StarVLA-GR00T}  & 20K & 83.0 & 59.4 & 18.8 & \textbf{100.0} & 65.3 \\
    \cmidrule(lr){2-8}

    & \multicolumn{7}{c}{\textbf{VL = Cosmos-Predict2-2B}} \\
    & \textbf{StarVLA-OFT} & 30K & 66.8 & 62.6 & 25.3 & 90.2 & 61.2 \\
    & \textbf{StarVLA-$\pi$} & 30K & 81.4 & 55.2 & 25.1 & 73.0 & 58.7 \\
    & \textbf{StarVLA-GR00T} & 30K & 80.4 & 65.4  & 20.0 & 80.6 & 61.6 \\

    \bottomrule
  \end{tabular}
  \end{adjustbox}
  \vspace{-0.3 em}
\end{table}

\begin{table}[ht!]
  \centering\footnotesize
  \caption{Detailed results on the SimplerEnv Google Robot benchmark. Numbers are officially reported unless marked with $*$, which denotes our reimplementation. We report StarVLA-OFT with Qwen3-VL-4B as a representative configuration due to the high evaluation cost on this platform.}
  \begin{tabular}{ l l c c c c c }
    \toprule
    \makecell[c]{Google\\Robot}
    & \multicolumn{1}{c}{Models}
    & \makecell[c]{Pick \\ Coke Can}
    & \makecell[c]{Move \\ Near}
    & \makecell[c]{Open/Close \\ Drawer}
    & \makecell[c]{Open Top Drawer \\ and Place Apple}
    & Avg \\
    \midrule
    \multirow{10}{*}{\makecell[l]{Visual\\Matching}}
      & RT-1~\cite{RT-1}      & 85.7 & 44.2 & \textbf{{73.0}} &  6.5 & 52.4 \\
      & RT-1-X~\cite{open_x_embodiment}  & 56.7 & 31.7 & 59.7 & 21.3 & 42.4 \\
      & RT-2-X~\cite{RT-2}  & 78.7 & 77.9 & 25.0 &  3.7 & 46.3 \\
      & OpenVLA~\cite{openvla}    & 18.0 & 56.3 & 63.0 &  0.0 & 34.3 \\
      & {CogACT}~\cite{cogact}      & {91.3} & {85.0} & {71.8} & {50.9} & 74.8 \\
      & SpatialVLA~\cite{spatialvla} & 86.0 & 77.9 & 57.4 & - & {75.1} \\
      & $\pi_0$~\cite{pi_0}     &  72.7  & 65.3 & 38.3 & - & 58.8  \\
      & $\pi_0$-FAST~\cite{pertsch2025fast}  & 75.3 & 67.5 & 42.9  & -  & 61.9 \\
      & GR00T N1.5$^*$~\cite{bjorck2025gr00t}  & 51.7 & 54.0 & 27.8 & 7.4 & 35.2 \\
      & Magma~\cite{magma}   & 83.7 & 65.4  & 56.0 &  6.4 & 52.9 \\
      \cmidrule(lr){2-7}
      & \textbf{StarVLA-OFT}   &  \textbf{95.3} & {75.0} & {68.8} & {66.1} & {76.0} \\
    \midrule
    \multirow{10}{*}{\makecell[l]{Variant\\Aggregation}}
      & RT-1~\cite{RT-1}      & {89.8} & 50.0 & 32.3 &  2.6 & 43.7 \\
      & RT-1-X~\cite{open_x_embodiment}  & 49.0 & 32.3 & 29.4 & 10.1 & 30.2 \\
      & RT-2-X~\cite{RT-2}  & 82.3 & 79.2 & 35.3 & 20.6 & 54.4 \\
      & OpenVLA~\cite{openvla}    & 60.8 & 67.7 & 28.8 &  0.0 & 39.3 \\
      & {CogACT}~\cite{cogact}      & {89.6} & {{80.8}} & 28.3 & {46.6} & {61.3} \\
      & SpatialVLA~\cite{spatialvla} & 88.0 & \textbf{82.5} & {41.8} & - & {70.7} \\
      & $\pi_0$~\cite{pi_0}     & 75.2 & 63.7 & 25.6 & - & 54.8 \\
      & $\pi_0$-FAST~\cite{pertsch2025fast}  & 77.6 & 68.2 & 31.3 & - & 59.0 \\
      & GR00T N1.5~\cite{bjorck2025gr00t}  & 69.3 & 68.7 & 35.8 & 4.0 & 44.5 \\
      & Magma~\cite{magma}  & 68.8 & 65.7  & {53.4} & 18.5 & 51.6 \\
      \cmidrule(lr){2-7}
      & \textbf{StarVLA-OFT} & {91.3} & {75.1} & {55.0} & {59.4} & {70.2} \\
      \bottomrule
  \end{tabular}
  \label{tab:simpler-google}
\end{table}

%% file: Sections/Benchmarks/A3-RoboCasa.tex
\subsection{Results on RoboCasa-GR1}


\paragraph{Training setup.}
We train the RoboCasa-GR1 baselines with distributed full-parameter fine-tuning on 16 A100 GPUs. Unless otherwise specified, the per-device batch size is 16 and training runs for up to 100K optimization steps. Checkpoints are saved every 10K steps, with periodic logging and evaluation during training. 
For the specialist setting, we use the official RoboCasa-GR1 tabletop release and train one model jointly across all 24 tasks from this benchmark only. This keeps the policy architecture fixed while treating RoboCasa as a multi-task humanoid-style manipulation suite rather than 24 separate single-task runs. 

\paragraph{Evaluation protocol.}
We follow the official RoboCasa-GR1 evaluation workflow and report average success rate over the 24 tasks. For the architecture comparison in this section, each model is evaluated with 50 rollouts per task. Table~\ref{tab:robocasa-results} further reports the task-level success rates for representative baselines and StarVLA variants.

\paragraph{Results.}
Table~\ref{tab:robocasa_summary} summarizes the average RoboCasa-GR1 performance for the single-benchmark setting. This benchmark is noticeably harder than LIBERO and SimplerEnv, and the choice of action head matters more: the discrete StarVLA-FAST baseline reaches 39.0\%, while the continuous-action variants improve to 43.9--48.8\%. Among the StarVLA variants, StarVLA-OFT performs best with a 48.8\% average success rate, slightly exceeding StarVLA-GR00T (47.8\%) and outperforming $\pi_{0.5}$ by 11.8 points. Detailed task-level results are reported in Table~\ref{tab:robocasa-results}. We defer cross-benchmark generalist results to Sec.~\ref{sec:all_in_one}.

\begin{table}[tbp]
\centering
\caption{Average success rate on RoboCasa-GR1 (24 tasks) under the single-benchmark training setting.}
\label{tab:robocasa_summary}
\small
\begin{tabular}{l c l c}
\toprule
\textbf{Method} & \textbf{Avg (\%)} & \textbf{Method} & \textbf{Avg (\%)} \\
\midrule
$\pi_{0.5}$~\cite{pi_05} & 37.0 & GR00T-N1.6~\cite{bjorck2025gr00t} & 47.6 \\
StarVLA-FAST & 39.0 & StarVLA-$\pi$ & 43.9 \\
StarVLA-GR00T & 47.8 & StarVLA-OFT & \textbf{48.8} \\
\bottomrule
\end{tabular}
\end{table}

\input{Sections/Tables/bench_robocasa}

%% file: Sections/Tables/bench_robocasa.tex
\begin{table}[htbp]
\centering
\caption{RoboCasa GR1 Tabletop Tasks Evaluation Results. A single model was trained for all 24 tasks. Results are reported over 50 rollouts per task (average success rate with 250 rollouts: 48.97\%).
}
\label{tab:robocasa-results}
\begin{adjustbox}{max width=\textwidth}
\begin{tabular}{lccccc}
\toprule
\textbf{Task} & \textbf{GR00T-N1.6} & \textbf{StarVLA-GR00T} & \textbf{StarVLA-$\pi$} & \textbf{StarVLA-OFT} & \textbf{StarVLA-FAST} \\
\midrule
PnPBottleToCabinetClose & \textbf{51.5} & 46.0 & 26.0 & \textbf{30.0} & 38.0 \\
PnPCanToDrawerClose & 13.0 & 80.0 & 62.0 & {76.0} & 44.0 \\
PnPCupToDrawerClose & 8.5 & 54.0 & 42.0 & \textbf{44.0} & \textbf{56.}0 \\
PnPMilkToMicrowaveClose & 14.0 & 48.0 & \textbf{50.0} & \textbf{44.0} & 44.0 \\
PnPPotatoToMicrowaveClose & 41.5 & 28.0 & \textbf{42.0} & \textbf{32.0} & 14.0 \\
PnPWineToCabinetClose & 16.5 & \textbf{46.0} & 32.0 & \textbf{36.0} & 14.0 \\
\addlinespace
PnPNovelFromCuttingboardToBasket & \textbf{58.0} & 48.0 & 40.0 & \textbf{50.0} & 54.0 \\
PnPNovelFromCuttingboardToCardboardbox & \textbf{46.5} & 40.0 & 46.0 & \textbf{40.0} & 42.0 \\
PnPNovelFromCuttingboardToPan & 68.5 & 68.0 & \textbf{70.0} & {70.0} & 58.0 \\
PnPNovelFromCuttingboardToPot & \textbf{65.0} & 52.0 & 40.0 & {54.0} & 58.0 \\
PnPNovelFromCuttingboardToTieredbasket & 46.5 & \textbf{56.0} & 44.0 & \textbf{38.0} & 40.0 \\
\addlinespace
PnPNovelFromPlacematToBasket & \textbf{58.5} & 42.0 & 44.0 & \textbf{32.0} & 36.0 \\
PnPNovelFromPlacematToBowl & 57.5 & \textbf{44.0} & 52.0 & \textbf{58.0} & 38.0 \\
PnPNovelFromPlacematToPlate & \textbf{63.0} & 48.0 & 50.0 & \textbf{52.0} & 42.0 \\
PnPNovelFromPlacematToTieredshelf & \textbf{28.5} & 18.0 & 28.0 & \textbf{24.0} & 18.0 \\
\addlinespace
PnPNovelFromPlateToBowl & 57.0 & \textbf{60.0} & 52.0 & \textbf{60.0} & 52.0 \\
PnPNovelFromPlateToCardboardbox & 43.5 & \textbf{50.0} & 40.0 & \textbf{50.0} & 30.0 \\
PnPNovelFromPlateToPan & 51.0 & \textbf{54.0} & 36.0 & \textbf{66.0} & 48.0 \\
PnPNovelFromPlateToPlate & 78.7 & \textbf{70.0} & 48.0 & \textbf{68.0} & 50.0 \\
\addlinespace
PnPNovelFromTrayToCardboardbox & \textbf{51.5} & 38.0 & 34.0 & {44.0} & 28.0 \\
PnPNovelFromTrayToPlate & \textbf{71.0} & 56.0 & 64.0 & \textbf{56.0} & 34.0 \\
PnPNovelFromTrayToPot & \textbf{64.5} & 50.0 & 44.0 & \textbf{62.0} & 46.0 \\
PnPNovelFromTrayToTieredbasket & \textbf{57.0} & 36.0 & 50.0 & \textbf{54.0} & 36.0 \\
PnPNovelFromTrayToTieredshelf & \textbf{31.5} & 16.0 & 28.0 & \textbf{30.0} & 16.0 \\
\midrule
\textbf{Average} & \textbf{47.6} & \textbf{47.8} & \textbf{43.9} & \textbf{48.8} & 39.0 \\
\bottomrule
\end{tabular}
\end{adjustbox}
\end{table}

%% file: Sections/Benchmarks/A3-RoboTwin.tex
\subsection{Results on Robotwin 2.0}

\paragraph{Training setup.}
We train the RoboTwin 2.0 baseline using distributed training with 48 A100 GPUs (via \texttt{accelerate} + DeepSpeed ZeRO-2). Unless otherwise specified, the per-device batch size is 4 and training runs for 150K optimization steps. Checkpoints are saved every 10K steps, with periodic logging and evaluation during training. For transparency and exact reproducibility (full command line, YAML configuration, and environment variables), we provide the complete training scripts under \texttt{examples/Robotwin/train\_files/}. We train RoboTwin 2.0 baselines on official clean and randomized datasets in LeRobot format: \url{https://huggingface.co/datasets/StarVLA/RoboTwin-Clean} and \url{https://huggingface.co/datasets/StarVLA/RoboTwin-Randomized}.


\paragraph{Evaluation protocol.}
We evaluate on the 50 tasks using the official RoboTwin 2.0 evaluation scripts and report success rate. 
We periodically evaluate checkpoints (every 10K steps by default) and report the earliest checkpoint that achieves the best average success rate.
For each suite, we run 50 tasks with 100 episodes per task under clean and randomized condition (10000 trials total) and report the mean success rate over all trials. To ensure reproducibility without modifying benchmark logic, we provide the complete evaluation scripts and launch instructions under \texttt{examples/Robotwin/eval\_files/}.

\paragraph{Results.}
Table~\ref{tab:robotwin_results} summarizes the RoboTwin baseline performance. Under Qwen3-VL-4B backbones, all four StarVLA variants achieve strong average success rates when trained as a single unified policy over 50 tasks, demonstrating that our end-to-end baseline pipeline (data \textrightarrow{} training \textrightarrow{} evaluation) is reliable and reproducible.

\begin{table}[tbp]
\centering
\caption{Detailed results on the RoboTwin 2.0 benchmark. We report different StarVLA model architecture on this platform.}
\label{tab:robotwin_results}
\small
\begin{tabular}{lcc|lcc}
\toprule
\textbf{Method} & \textbf{Clean} & \textbf{Random} & \textbf{Method} & \textbf{Clean} & \textbf{Random} \\
\midrule
$\pi_0$~\cite{pi_0}         & 65.9  & 58.4 & $\pi_{0.5}$~\cite{pi_05}     & 82.7  & 76.8 \\
X-VLA~\cite{zheng2025x}   & 72.9 & 72.8 & Lingbot-VLA~\cite{wu2026pragmatic}     & 88.6  & 86.7 \\
StarVLA-FAST    & 72.5  & 83.2 & StarVLA-OFT     & 88.2  & 88.3 \\
StarVLA-GR00T   & 88.0  & 88.5 & StarVLA-$\pi$   & 88.1  & 88.8 \\
\bottomrule
\end{tabular}
\end{table}

%% file: Sections/A5-CoTraining.tex
\section{Multimodal Co-Training Examples}
\label{sec:cotrain}

Beyond single-benchmark supervised fine-tuning, StarVLA natively supports \emph{multimodal co-training}, in which the VLM backbone is jointly optimized on both robot action data and auxiliary vision-language tasks (e.g., spatial grounding, visual question answering, and captioning).
The motivation is twofold: (i) action-only fine-tuning can rapidly degrade the pre-trained multimodal representations, undermining instruction comprehension and spatial reasoning; and (ii) co-training with carefully curated auxiliary data can align the optimization dynamics of perception and control, leading to better-performing policies.

When a pre-trained VLM is fine-tuned exclusively on action prediction, it tends to ``forget'' pre-trained visual and linguistic capabilities within thousands of steps.
This manifests as degraded object grounding, instruction following, and scene understanding, all of which are prerequisites for robust manipulation.
Co-training with multimodal grounding data counteracts this forgetting by maintaining the gradient flow through perception-relevant pathways.

\subsection{Experimental Setup}

\mypara{Co-training setup.}
StarVLA provides built-in support for mixing heterogeneous data sources during training.
Users can specify arbitrary combinations of action datasets and VLM-style QA datasets in a single configuration file; the framework handles tokenization, loss masking, and gradient accumulation transparently across data types.
This makes it straightforward to reproduce co-training recipes, for instance, mixing OXE action data with RefCOCO spatial grounding or LLaVA-style visual QA data, without modifying the training loop.

\mypara{Evaluation and baselines.} To illustrate the effect, we summarize a spatially guided co-training study built on the StarVLA codebase~\citep{ye2026st4vla}. This study compares three training strategies: (1)~\emph{Vanilla VLA}, which fine-tunes only on action data, (2)~\emph{Vanilla co-training VLA}, which jointly optimizes on spatial grounding and action data, and (3)~\emph{Spatially guided training VLA}, which additionally incorporates spatial pre-training and spatial prompting during co-training.

\subsection{Main Results for Multimodal Co-training}

\begin{figure}[ht]
    \centering
    \includegraphics[width=1.\textwidth]{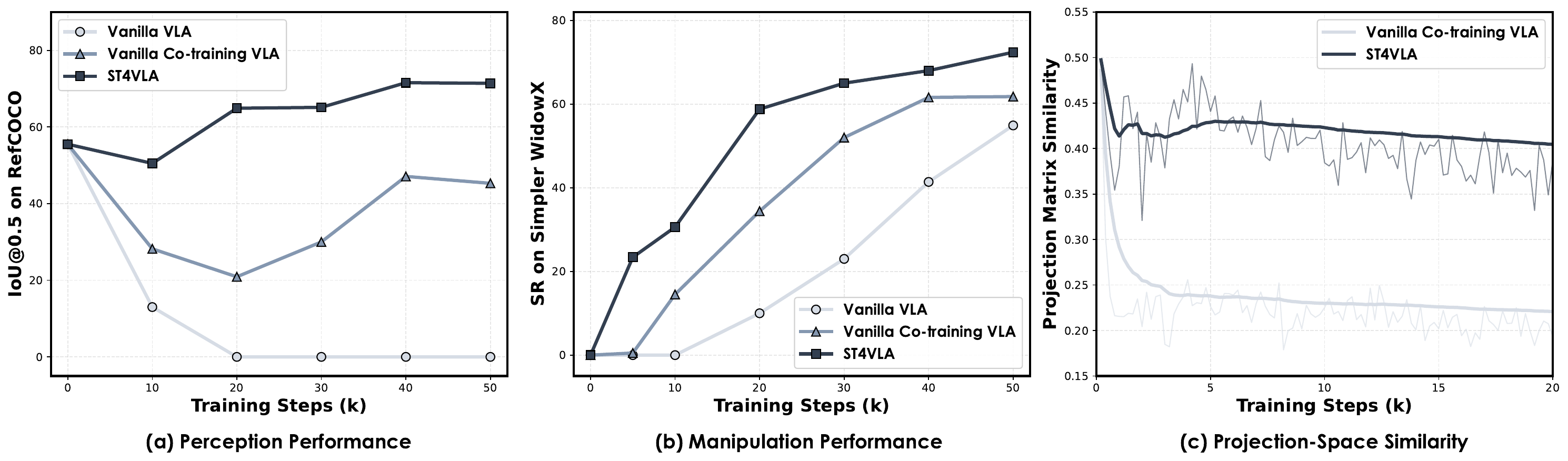}
    \caption{Perception--action co-optimization dynamics under different co-training strategies (reproduced from a StarVLA-based spatially guided co-training study ST4VLA~\citep{ye2026st4vla}).
    From left to right: (a) spatial grounding performance (IoU@0.5 on RefCOCO-g); (b) manipulation success rate (WidowX); (c) gradient subspace alignment (PSS) between spatial grounding and action objectives under vanilla co-training vs.\ spatially guided co-training.}
    \label{fig:cotrain_curves}
\end{figure}

Figure~\ref{fig:cotrain_curves} visualizes the interaction between spatial perception (measured by IoU@0.5 on RefCOCO-g) and manipulation performance (WidowX success rate) across training steps.
Vanilla VLA suffers rapid perception degradation: RefCOCO-g performance drops to near-random levels within 20K steps.
Vanilla co-training partially preserves perception but exhibits unstable oscillations.
The spatially guided StarVLA~(ST4VLA~\citep{ye2026st4vla}) variant achieves the best balance, maintaining ${\sim}70\%$ of original grounding performance while reaching strong manipulation success.

Table~\ref{tab:cotrain_probing} further quantifies the impact of co-training on multimodal understanding, spatial grounding, and robotic manipulation.
Compared to the vanilla VLA, vanilla co-training already improves manipulation performance (+4.1\% Google Robot VM, +6.4\% WidowX) while recovering multimodal capabilities.  
The spatially guided StarVLA variant pushes the results further, achieving 84.6\%/75.9\% on Google Robot VM/VA and 73.2\% on WidowX, while simultaneously preserving strong spatial grounding (71.2 IoU@0.5 on RefCOCO-g).

\begin{table}[ht]
  \centering
  \caption{Effect of co-training strategies on multimodal understanding, spatial grounding, and robotic manipulation (from a StarVLA-based spatially guided co-training study~\cite{ye2026st4vla}).}
  \label{tab:cotrain_probing}
  \begin{adjustbox}{width=\linewidth}
  \begin{tabular}{l cccc cc cc}
    \toprule
    & \multicolumn{4}{c}{\textbf{Multi-modal Understanding}}
    & \multicolumn{2}{c}{\textbf{Spatial Grounding}} 
    & \multicolumn{2}{c}{\textbf{Robotic Manipulation}} 
    \\
    \cmidrule(lr){2-5}
    \cmidrule(lr){6-7}
    \cmidrule(lr){8-9}
    \textbf{Training Strategy} 
      & \makecell[c]{MME} 
      & \makecell[c]{MMVet} 
      & \makecell[c]{TextVQA} 
      & \makecell[c]{POPE\\Acc} 
      & \makecell[c]{RefCOCO-g\\IoU@0.5} 
      & \makecell[c]{RoboRefIt\\Acc@0.5} 
      & \makecell[c]{Google Robot\\VM / VA} 
      & \makecell[c]{WidowX\\VM} \\
    \midrule
    Vanilla VLA & -- & -- & -- & -- & -- & -- & 66.1 / 63.5 & 54.7 \\
    + Co-training & 1106 & 19.2 & 20.5 & 78.0 & 47.1 & 66.7 & 70.2 / 66.5 & 61.1 \\
    + Spatially guided & 1374 & 23.0 & 28.4 & 84.6 & 68.1 & 72.5 & 78.8 / 70.0 & 67.4 \\
    + Spatially pretrained & 1411 & 23.3 & 28.6 & 86.2 & 71.2 & 74.3 & 84.6 / 75.9 & 73.2 \\
    \bottomrule
  \end{tabular}
  \end{adjustbox}
\end{table}

\mypara{Takeaways.} These results demonstrate that StarVLA's co-training infrastructure enables significant gains over action-only fine-tuning.
By preserving multimodal understanding during policy learning, co-training yields more generalizable agents.
For a comprehensive treatment of spatially guided co-training, including the full training recipe, gradient alignment analysis, and extensive real-world experiments, we refer readers to the ST4VLA~\cite{ye2026st4vla}, a study paper based on StarVLA.

%% file: Sections/A6-All-Bench-in-One.tex
\section{Cross-Benchmark Training Examples}
\label{sec:all_in_one}
Building on the benchmark-wise specialist baselines in Sec.~\ref{sec:Basic_experiments}, we next evaluate a stricter setting for embodied generalization: \emph{one model jointly trained across benchmarks and robot embodiments}. StarVLA natively supports co-training on heterogeneous datasets under a unified framework, which makes this all-in-one setting a natural case study for generalist VLA training.

\begin{table}[t]
\centering
\caption{\textbf{Performance comparison between generalist and specialist settings.} Specialist represents multiple models trained separately on each benchmark-specific dataset, while Generalist represents a single model jointly trained across all datasets.
}
\label{tab:general-results}

\begin{adjustbox}{max width=\linewidth}
\begin{tabular}{l l c c c c c c c c c c c c}
\toprule
\multirow{2}{*}{\textbf{Settings}} & \multirow{2}{*}{\textbf{Method}}
& \multicolumn{5}{c}{LIBERO}
& \multicolumn{3}{c}{SimplerEnv}
& \multicolumn{3}{c}{RoboTwin 2.0}
& \multirow{1}{*}{RoboCasa-GR1} \\
\cmidrule(lr){3-7} \cmidrule(lr){8-10} \cmidrule(lr){11-13} \cmidrule(lr){14-14}
& & Spatial & Object & Goal & Long & avg
& WidowX & Google VA & Google VM
& clean & clean$^*$ & random$^*$
& (avg of 24 tasks) \\
\midrule

\multirow{5}{*}{\textbf{Specialist}}
& $\pi_{0.5}$   & 98.8 & 98.2 & 98.0 & 92.4 & 96.9 & 46.9  & 68.4 & 72.7 &  \textbf{60.2} & 82.7 & 76.8 & 37.0 \\
& GR00T-N1.6    & 97.5 & 98.5 & 97.5 & 94.4 & 94.1 & 67.8 & 41.5 & 35.2 & --   & --   & --   & 47.6 \\
\cmidrule(lr){2-14}

& StarVLA-$\pi$     & 98.0 & 99.2 & 98.2 & 93.6 & 98.1 & 65.9 & 72.8 & 76.6 & 50.8 & 88.1 & \textbf{88.8} & 48.9 \\
& StarVLA-GR00T  & 98.9 & 99.6 & 98.4 & \textbf{95.3} & 98.7 & 65.3 & 70.7 & 75.3 & 48.8 & 88.0 & 88.5 & 52.8 \\
& StarVLA-OFT    & \textbf{99.0} & \textbf{99.8} & 98.5 & 94.1 & \textbf{98.8} & 64.6 & 70.2 & 76.0 & 53.4 & 88.2 & 88.3 & 53.8 \\
\midrule

\textbf{Generalist}
& StarVLA  & 98.7 & 99.7 & \textbf{98.6} & 94.2 & 97.8 & \textbf{70.2} & \textbf{73.8} & \textbf{79.3} & --   & \textbf{88.7} & 87.8 & \textbf{57.3} \\
\bottomrule
\end{tabular}
\end{adjustbox}
\end{table}

\mypara{Existing evaluation patterns.} The Embodied AI community shares a common ambition: to develop a generalist agent that can seamlessly operate across diverse tasks, environments, and robots. In practice, however, the research landscape remains fragmented. Many state-of-the-art systems are tuned for specific benchmarks, and their performance can drop substantially when transferred to different environments or embodiments. This makes it difficult to measure true generalization ability.

\subsection{Experimental Setups}

\mypara{Training setup.}
In this setting, we jointly train one model on the merged training sets from LIBERO, SimplerEnv, RoboTwin 2.0, and RoboCasa-GR1, and then directly evaluate on each benchmark under its official protocol. No additional benchmark-specific fine-tuning is applied. We set the learning rate to $1\times10^{-4}$, the total batch size to 256, and train jointly on the merged benchmark datasets. To handle action-space differences across embodiments, we avoid task-specific action heads and apply a simple unified padding strategy that expands lower-DoF actions to a shared 32-dimensional action vector.

\mypara{Evaluation protocol as a generalist.} A practical way to test generalization is to require one model to handle diverse benchmarks simultaneously. Following this principle, we evaluate StarVLA under a unified multi-benchmark setting, where a single policy is trained once and evaluated across suites without benchmark-specific fine-tuning.

\mypara{Baselines.}
To further demonstrate the effectiveness of our method and the proposed setting, we report both specialist results, where models are trained only on individual datasets, and results from the generalist training setting. In addition to comparing with our model, we also evaluate several state-of-the-art methods, such as $\pi_{0.5}$ and GR00T-N1.6.

\subsection{Main Results as a Generalist}

As shown in Table~\ref{tab:general-results}, we compare our generalist model (jointly trained across datasets) with specialist models trained per benchmark. The generalist model remains competitive across most benchmarks and improves RoboCasa-GR1 from the best specialist average of 48.8\% to 57.3\% on the 24-task average. These results support the feasibility of a single policy that transfers across tasks and embodiments under a unified training/evaluation setting.

\mypara{Takeaways}
This section focuses on a direct capability demonstration rather than ablation analysis: StarVLA can jointly train on heterogeneous, cross-embodiment benchmark datasets and produce a single model that remains competitive across diverse evaluation suites. We view this as evidence that all-in-one multi-benchmark training is a practical path toward large-scale cross-embodiment pretraining for future generalist VLA systems.

%% file: Sections/A7-Efficiency-Report.tex
\section{Computation Efficiency}

This section reports the training efficiency of StarVLA using the public profiling measurements collected in issue \url{https://github.com/starVLA/starVLA/issues/158}.
Our goal is to provide actionable scaling guidance for practitioners, while keeping the reported metrics aligned with common distributed-training bottlenecks (compute and communication).

\begin{table}[h]
	\centering\small
	\caption{Single-node training efficiency (8\,$\times$\,A100). Sample throughput is derived from the measured time per 100K steps and the global batch size.}
	\label{tab:eff_single_node}
	\begin{tabular}{c c c c c c}
		\toprule
		\textbf{Per-GPU batch} & \textbf{Global batch} & \textbf{Time / 100K steps} & \textbf{Seconds / step} & \textbf{Samples / s} & \textbf{GPU util} \\
		\midrule
		2  & 16  & 19:32:17 & 0.703 & 22.7  & 74\% \\
		4  & 32  & 24:35:59 & 0.886 & 36.1  & 89\% \\
		8  & 64  & 31:25:38 & 1.131 & 56.6  & 92\% \\
		16 & 128 & 49:15:53 & 1.774 & 72.2  & 91\% \\
		24 & 192 & 66:47:02 & 2.404 & 79.9  & 96\% \\
		\bottomrule
	\end{tabular}
\end{table}

\begin{figure}[th]
	\centering
    \includegraphics[width=\linewidth]{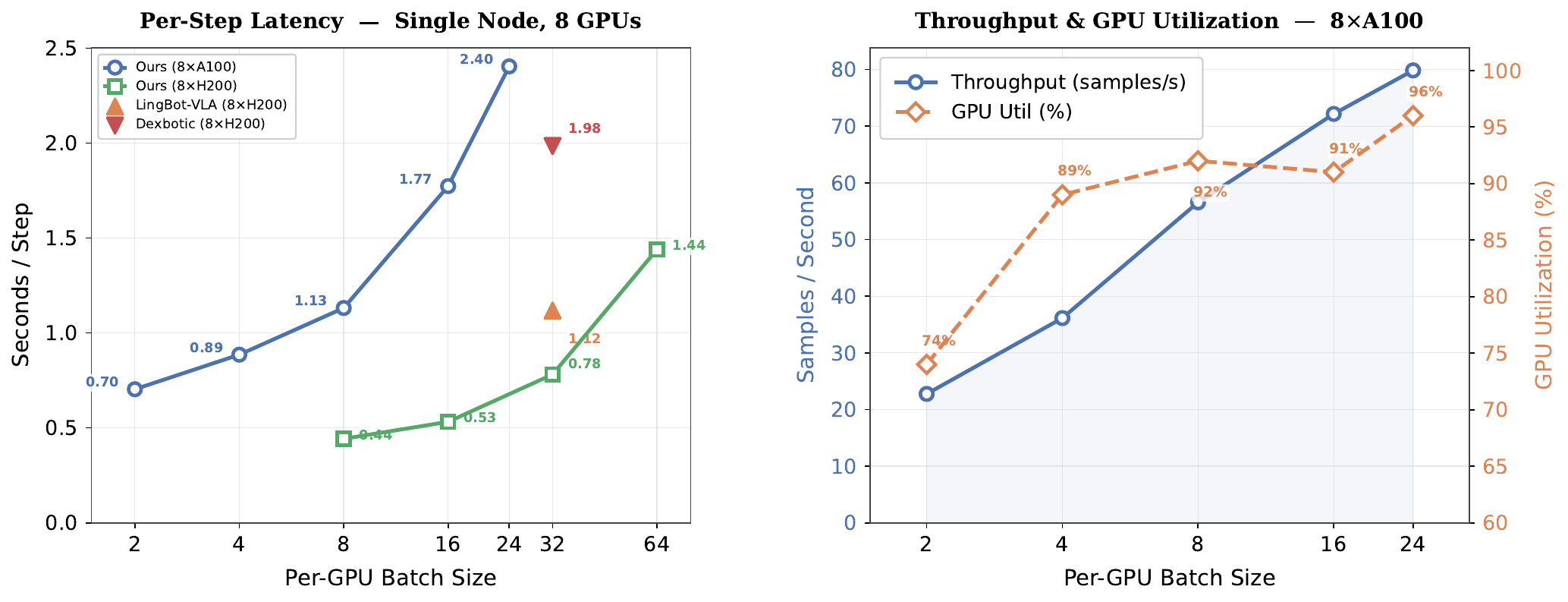}%
	\caption{\textbf{Per-step latency and throughput on a single 8-GPU node.} Left: step latency as a function of per-GPU batch size for our method on A100 and H200, compared with LingBot-VLA and Dexbotic (both on 8×H200). Right: training throughput and GPU utilization on 8×A100 across batch sizes.}
	\label{fig:eff_single_node}
\end{figure}

\begin{table}[t]
	\centering
	\caption{Multi-node training efficiency (per-GPU batch = 8). ``Ideal'' scaling assumes linear growth of samples/s from the 8-GPU baseline.}
	\label{tab:eff_multi_node}
    \small
	\begin{tabular}{c c c c c c}
		\toprule
		\textbf{\# GPUs} & \textbf{Global batch} & \textbf{Time / 100K steps} & \textbf{Seconds / step} & \textbf{Samples / s} & \textbf{Scaling eff.} \\
		\midrule
		8   & 64   & 20:25:48 & 0.735 & 87.0   & 100\% \\
		16  & 128  & 23:36:00 & 0.850 & 150.7  & 86.7\% \\
		32  & 256  & 24:58:45 & 0.899 & 284.7  & 81.9\% \\
		64  & 512  & 25:40:59 & 0.925 & 553.8  & 79.6\% \\
		128 & 1024 & 25:35:26 & 0.921 & 1111.5 & 79.9\% \\
		256 & 2048 & 25:51:41 & 0.931 & 2200.0 & 79.1\% \\
		\bottomrule
	\end{tabular}
\end{table}

\begin{figure}[h]
	\centering
    \includegraphics[width=\linewidth]{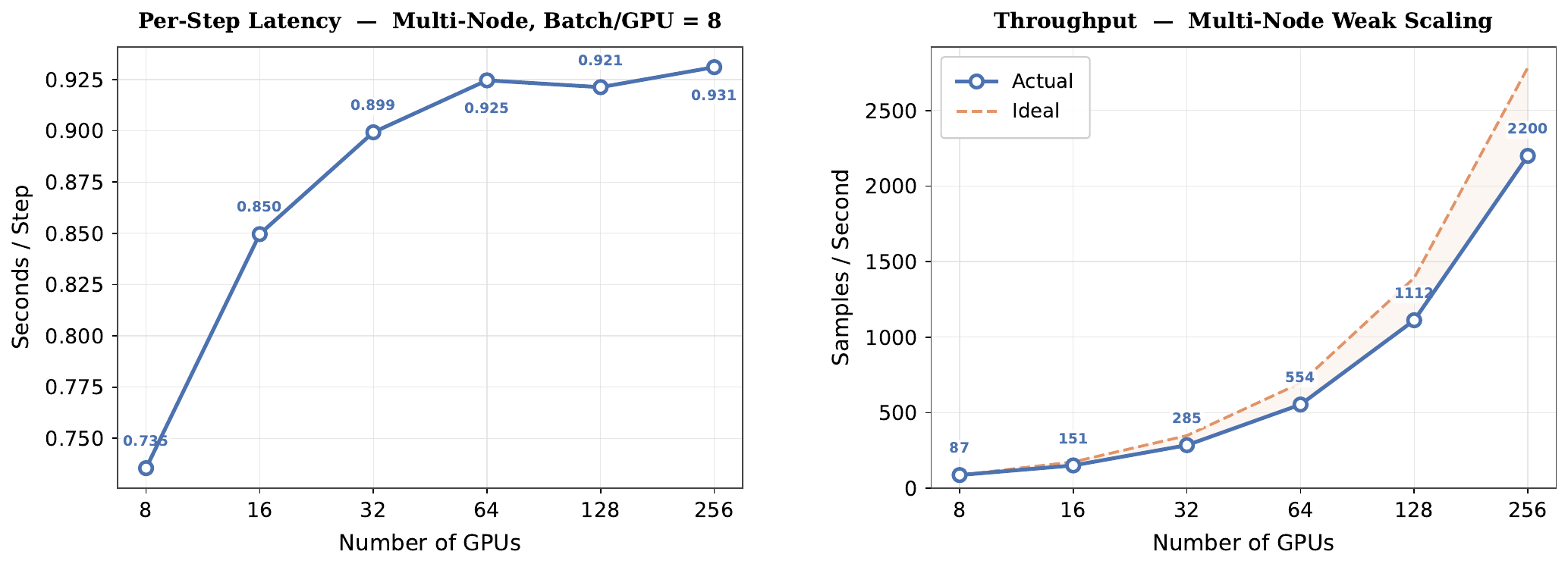}
	\caption{\textbf{Multi-node scaling efficiency.} Left: per-step latency rises noticeably from 8 to 32 GPUs due to inter-node communication overhead, then plateaus between 64 and 256 GPUs. Right: measured sample throughput versus ideal linear scaling; parallel efficiency stabilizes around 79--80\% beyond 32 GPUs.}
	\label{fig:eff_multi_node}
\end{figure}

\paragraph{Experimental setup.}
Unless otherwise specified, the measurements use StarVLA-GR00T with a Qwen3-VL-4B backbone trained on the RoboCasa-GR1 dataset on A100 80GB GPUs.
We report wall-clock time per 100K optimization steps, which includes distributed communication and system overhead.

\paragraph{Efficiency metrics.}
We distinguish two throughput notions:
(i) \emph{step throughput} (lower seconds/step is better), and
(ii) \emph{sample throughput} (higher samples/s is better), where samples/s is computed as $\text{global batch} / (\text{seconds per step})$.
This distinction is important because distributed scaling often decreases step throughput (due to synchronization) while increasing sample throughput (due to larger global batch).

\subsection{Single-Node Training Efficiency}
Table~\ref{tab:eff_single_node} summarizes a single-node sweep that varies the per-GPU batch size.
We omit derived ``24-hour'' projections and focus on directly measured quantities and the implied sample throughput.

Figure~\ref{fig:eff_single_node} visualizes the main trade-off.
Smaller per-GPU batches yield faster steps (e.g., 0.703\,s/step at batch~2 vs.\ 2.404\,s/step at batch~24), while larger per-GPU batches improve sample throughput (from 22.7 to 79.9\,samples/s) at the cost of sharply increased step latency.

\subsection{Multi-Node Scaling Efficiency}
We next fix per-GPU batch size to 8 and scale the number of GPUs.
As shown in Table~\ref{tab:eff_multi_node}, the time per step rises from 0.735\,s (8~GPUs) to 0.899\,s (32~GPUs) due to inter-node communication overhead, then plateaus at $\sim$0.93\,s up to 256~GPUs.
Despite this overhead, sample throughput scales from 87.0 to 2200.0\,samples/s, which is the relevant metric when the training objective is to process a fixed amount of data quickly.

Figure~\ref{fig:eff_multi_node} plots both step latency and sample throughput against GPU count, together with the ideal linear reference line.
The results highlight a practical guideline: scaling out is most beneficial for data-volume-driven training, while fixed-step training does not become faster with more GPUs.

\mypara{Takeaways.}
First, inter-node communication introduces a one-time latency overhead (0.735$\to$0.93\,s/step), but sample throughput still scales near-linearly via larger global batch.
Second, on a single node, a moderate per-GPU batch (e.g., 8) often provides the best balance between step latency and GPU utilization; very large batches (e.g., 24) maximize utilization (96\%) but inflate step latency by $3.4\times$.
Third, for large-scale training, once the system scales beyond 8 nodes (64~GPUs), the communication burden no longer grows further, maintaining a stable scaling efficiency of 79--80\%. This indicates that practitioners can confidently scale to hundreds of GPUs without incurring additional parallel efficiency degradation.

%% file: Sections/Tables/author_list.tex
\section{Authors and Contributors for StarVLA v1.0}
\label{Authors}

StarVLA thrives on the synergy between its dedicated core team and a vibrant open-source community. To accurately reflect the nature of involvement, we list contributors in two categories: \textbf{Authors} and \textbf{Community Contributors}. We extend our deepest gratitude to everyone who has helped shape and scale StarVLA.

\medskip

\paragraph{Authors.}

Jinhui Ye, Ning Gao, Yilun Chen\textsuperscript{$\dagger$}, Weiyu Guo, Zixuan Wang, Yuxing Chen, Fangjing Wang, Senqiao Yang, Chengyao Wang, Yuqi Liu, Meng Chu, Changsheng Lu, Pengguang Chen, Shu Liu\textsuperscript{$\dagger$}, Jiaya Jia\textsuperscript{$\dagger$$*$}

\paragraph{Community Contributors.}
Junqiu Yu, Shuang Zeng, Shijie Lian, Hanwen Wan, Changjiu Zhang, Zhijie Song, Mingsheng Li, Qiuyue Wang, Sicheng Xie, Jinliang Zheng,
Deyu Zhou, Jiaming Zhou, Lu Dai, Xiaorui Zhao

\paragraph{Contributor Policy.}
\emph{Authors} constitute the core team of StarVLA. This group is responsible for continuously iterating on core features, maintaining the foundational framework, and providing ongoing, long-term support for the project. Researchers and developers who are interested in making sustained, structural contributions and wish to join the core author team are highly encouraged to contact us.
\emph{Community Contributors} are the vital force behind the project's broader ecosystem. We continuously receive invaluable support from the open-source community—ranging from new feature implementations (pull requests) and bug fixes to constructive feedback. We deeply appreciate these efforts, which allow StarVLA to evolve rapidly. The full and actively updated contributor history is maintained at \href{https://starvla.github.io/contributors/}{starvla.github.io/contributors}.

{\small \textsuperscript{$\dagger$}Corresponding authors. $*$ Von Neumann Institute, HKUST}